\documentclass[conference]{IEEEtran}
\IEEEoverridecommandlockouts
\usepackage{cite}
\usepackage{amsmath,amssymb,amsfonts}
\usepackage{algorithmic}
\usepackage{graphicx}
\usepackage{textcomp}
\usepackage{xcolor}
\usepackage{booktabs}
\usepackage{stfloats}
\usepackage{etoolbox}
\usepackage{geometry}
\usepackage{caption}

\def\BibTeX{{\rm B\kern-.05em{\sc i\kern-.025em b}\kern-.08em
    T\kern-.1667em\lower.7ex\hbox{E}\kern-.125emX}}

\title{A Physics-informed End-to-End Occupancy Framework for Motion Planning of Autonomous Vehicles}

\author{
\IEEEauthorblockN{1\textsuperscript{st} Shuqi Shen}
\IEEEauthorblockA{\textit{The Hong Kong University of} \\
\textit{Science and Technology (Guangzhou)}\\
Guangzhou, China \\
u202141021@xs.ustb.edu.cn}
\and
\IEEEauthorblockN{2\textsuperscript{nd} Junjie Yang}
\IEEEauthorblockA{\textit{The Hong Kong University of} \\
\textit{Science and Technology (Guangzhou)}\\
Guangzhou, China \\
jyang512@connect.hkust-gz.edu.cn}

\and
\IEEEauthorblockN{3\textsuperscript{rd} Hongliang Lu*}
\IEEEauthorblockA{\textit{The Hong Kong University of} \\
\textit{Science and Technology (Guangzhou)}\\
Guangzhou, China \\
hlu592@connect.hkust-gz.edu.cn\\
*Corresponding author}
\and
\IEEEauthorblockN{4\textsuperscript{th} Hui Zhong}
\IEEEauthorblockA{\textit{The Hong Kong University of} \\
\textit{Science and Technology (Guangzhou)}\\
Guangzhou, China \\
hzhong638@connect.hkust-gz.edu.cn}
\and
\IEEEauthorblockN{5\textsuperscript{th} Qiming Zhang}
\IEEEauthorblockA{\textit{The Hong Kong University of} \\
\textit{Science and Technology (Guangzhou)}\\
Guangzhou, China \\
qzhang255@connect.hkust-gz.edu.cn\\
}

\and
\IEEEauthorblockN{6\textsuperscript{th} Xinhu Zheng*}
\IEEEauthorblockA{\textit{The Hong Kong University of} \\
\textit{Science and Technology (Guangzhou)}\\
Guangzhou, China \\
xinhuzheng@hkust-gz.edu.cn\\
*Corresponding author}

\thanks{Corresponding authors: Hongliang Lu and Xinhu Zheng.}
}
\begin{document}

\maketitle

\begin{abstract}
Accurate and interpretable motion planning is essential for autonomous vehicles (AVs) navigating complex and uncertain environments. While recent end-to-end occupancy prediction methods have improved environmental understanding, they typically lack explicit physical constraints, limiting safety and generalization. In this paper, we propose a unified end-to-end framework that integrates verifiable physical rules into the occupancy learning process. Specifically, we embed artificial potential fields (APF) as physics-informed guidance during network training to ensure that predicted occupancy maps are both data-efficient and physically plausible. Our architecture combines convolutional and recurrent neural networks to capture spatial and temporal dependencies while preserving model flexibility. Experimental results demonstrate that our method improves task completion rate, safety margins, and planning efficiency across diverse driving scenarios, confirming its potential for reliable deployment in real-world AV systems.
\end{abstract}


\section{Introduction}


The recent rapid development of autonomous vehicles (AVs) worldwide has garnered widespread interest due to their promise to improve transportation and enhance road safety. 
AVs must possess accurate perception, prediction, and decision-making capabilities in complex traffic environments, where motion planning plays a crucial role.
Motion planning needs to generate a feasible, safe, and efficient trajectory that guides an AV from its current location to a desired destination within a known or partially known environment \cite{magdici2016fail}. 
A robust motion planning system typically accounts for spatial constraints and dynamic entities to ensure that the resulting trajectory is both physically executable and reasonable \cite{jiang2020dynamic}.
For motion planning, it is critical to represent entities in a reliable and effective manner, including their spatial locations, and dynamic movements.

In recent years, an approach called \textbf{occupancy} has attracted wide attention in representing entities effectively \cite{elfes2013occupancy}. Occupancy reflects whether each spatial location in the environment is occupied by other entities, which can help planning algorithms tell reachable areas, and avoid other entities \cite{genevois2023probabilistic}.
Classic occupancy methods mainly rely on handcrafted rules, such as probabilistic methods, for transforming sensor data to an occupancy representation \cite{kaufman2016bayesian}. For example, \cite{robbiano2020bayesian} proposed the Occupancy Grid Mapping approach, which combines LiDAR distance measurements with probabilistic update mechanisms to effectively represent occupancy states in static environments. However, this approach lacks the capacity to model sensor noise and struggles to adapt to complex dynamic environments \cite{sanchez2024transitional, nuss2018random}.
To address this, end-to-end occupancy has emerged. Instead of relying on explicit rules or handcrafted features \cite{bhattacharyya2021hybrid}, end-to-end occupancy uses artificial neural networks to directly learn occupancy from perception inputs, effectively reducing the impact of sensor noise \cite{schreiber2022multi}. 
For example, \cite{hoermann2018dynamic} employed convolutional neural network based models to predict dynamic occupancy probabilities from sensor-generated occupancy grid maps. \cite{jeon2018traffic} further combined convolutional neural network with Long Short-Term Memory(LSTM) architectures to achieve high-precision temporal predictions of dynamic occupancy in complex urban traffic scenarios.
However, this approach\newgeometry{top=0.75in, bottom=0.8in, left=0.75in, right=0.75in} may produce physically infeasible results, as they do not explicitly incorporate physical constraints during the learning process and rely entirely on data-driven learning \cite{karle2023mixnet}. 
To address this issue, some studies incorporate physics-informed approaches into the decision-making process. For example, \cite{zhou2024enhance} introduced a physics-informed safety controller to enhance physical feasibility and behavioral safety during execution. This approach often suffers from low computational efficiency and limited real-time performance, primarily because it depends on multiple separate modules and does not incorporate physical information into the network’s learning process.

In this paper, we propose a unified end-to-end occupancy framework that incorporates physical information into the learning process to support a reliable motion planner, while improving overall efficiency through integrated learning and planning.
To validate our framework, we incorporate the Artificial Potential Field (APF) method to provide physical guidance during learning and use it as a baseline for comparison.
It is worth noting that our framework allows for flexible physical considerations, enabling adaptation to different task requirements or physical priors.

This paper is organized as follows. Section II reviews the related work on occupancy modeling and physics-informed learning, emphasizing their limitations in dynamic and uncertain environments. Section III presents our proposed end-to-end occupancy framework, detailing the network architecture and the integration of physical constraints via artificial potential fields. Section IV reports the experimental results, including training behavior, scenario-based analysis, and a comprehensive evaluation across diverse driving conditions. Finally, Section V concludes the paper and outlines directions for future research.

\section{Related Work}

This section reviews previous research on motion planning and occupancy modeling. 
We begin by discussing rule-based occupancy representation approaches and then focus on the emerging trend of learning-based methods that leverage neural networks to directly infer occupancy states from perception data, highlighting their advantages in end-to-end representation learning.
Finally, we further focus on network learning methods that integrate physical rules, emphasizing significant progress in enhancing model controllability and safety. These methods have shown excellent adaptability and predictive capabilities, especially in high-dynamic environments like autonomous driving.

\subsection{Occupancy}
Occupancy Grid Mapping (OGM) is one of the earliest spatial modeling techniques adopted in autonomous driving systems \cite{suganuma2010robust}. Traditional OGM methods typically rely on hand-crafted rules and sensor models—such as range measurements from LiDAR or cameras—together with Bayesian filtering to infer the occupancy state of the environment. To improve modeling precision, efforts have been made to capture correlations between individual grid cells. For instance, \cite{elfes2013occupancy} introduced a Bayesian framework that explicitly models grid-to-grid correlations, thereby enhancing mapping accuracy. Similarly, \cite{dhiman2014modern} employed a forward sensor model with the Expectation-Maximization algorithm to better reflect spatial dependencies, which resulted in improved accuracy and computational efficiency.

Despite these advances, rule-based OGM methods remain limited in dynamic environments and noisy conditions due to their reliance on fixed models and simplified assumptions.

To overcome these limitations, recent work has turned to deep learning for more expressive and adaptive occupancy modeling. For example, SSCNet \cite{song2017semantic} integrates semantic perception with occupancy estimation, allowing the network to predict spatial occupancy and semantic labels with high accuracy jointly. However, the absence of physical priors in its learning process limits its interpretability. More recently, OFMPNet \cite{murhij2024ofmpnet} introduced a hybrid architecture combining Transformers, attention mechanisms, and convolutional units to improve prediction accuracy further. While this model achieves lower error rates, it still lacks explicit modeling of individual entities—such as vehicles and pedestrians—and does not leverage semantic or physical rules for guidance.

Taken together, the evolution from rule-based OGM to deep learning-based methods has significantly advanced occupancy modeling by improving generalization and performance in complex scenes. Nevertheless, the lack of physical rules in current learning paradigms remains a critical bottleneck. This gap not only hampers interpretability but also raises safety concerns, especially in safety-critical applications like autonomous driving. Therefore, future research should focus on integrating physics-informed priors with data-driven learning to achieve more robust, interpretable, and trustworthy occupancy models.

\subsection{Physics-informed} 
To overcome the limitations of purely data-driven methods in interpretability and controllability, recent research has turned to the integration of physical rules into deep learning-based occupancy modeling. These hybrid approaches typically incorporate physics either through post-processing modules or by introducing auxiliary constraints at the output stage, to improve the executability of predicted trajectories and occupancy states. While such strategies have shown practical value, their physical modeling often remains loosely coupled with the learning process. As a result, a unified, end-to-end differentiable framework that embeds physical rules directly into the learning architecture is still lacking.

Several recent works have made initial attempts toward this goal. For example, \cite{liao2024physics} proposed a two-stage trajectory prediction framework that employs a wavelet reconstruction network and imposes physical rules using a simplified kinematic bicycle model. Despite its strong performance in complex scenarios, the physical rule is implemented as an external enhancement module rather than being seamlessly integrated into the network itself. Similarly, \cite{liu2023occupancy} introduced a Transformer-based occupancy prediction framework, where the outputs are mapped to the Frenet coordinate system and refined via optimization. However, the physical constraints are still applied only during the post-processing stage, with no influence on the internal learning process. In another effort, \cite{mahjourian2022occupancy} developed a model that jointly predicts occupancy and flows over multiple time steps using an EfficientDet-based decoder. While it incorporates temporal consistency through velocity-based reasoning during learning process, the physical information appears only as flow-level constraints and is not fully embedded in the network’s structure.

These examples reflect a broader trend: although there is growing interest in incorporating physical rules into learning-based occupancy models, current approaches often rely on loosely attached or indirect mechanisms. As a result, they fall short of providing robust interpretability and controllability, particularly under real-world deployment conditions.

To address this issue, we propose an end-to-end framework that structurally embeds physical rules into both the learning and inference stages of neural networks. This approach aims to ensure that physical consistency is not treated as a post hoc correction but is instead learned jointly with spatial occupancy, thereby enhancing the model’s transparency, reliability, and safety in autonomous driving applications.

\section{Methodology}
This section presents methods for integrating physical rule constraints into occupancy map prediction frameworks. We first introduce the architecture design of the end-to-end network, followed by an explanation of how incorporating physical constraints enhances the model’s physical interpretability. In this work, Artificial Potential Fields (APF) are employed as the physical constraint module, which can be replaced with alternative formulations if needed.

\begin{figure*}[t]
    \centering
    \includegraphics[width=0.75\textwidth]{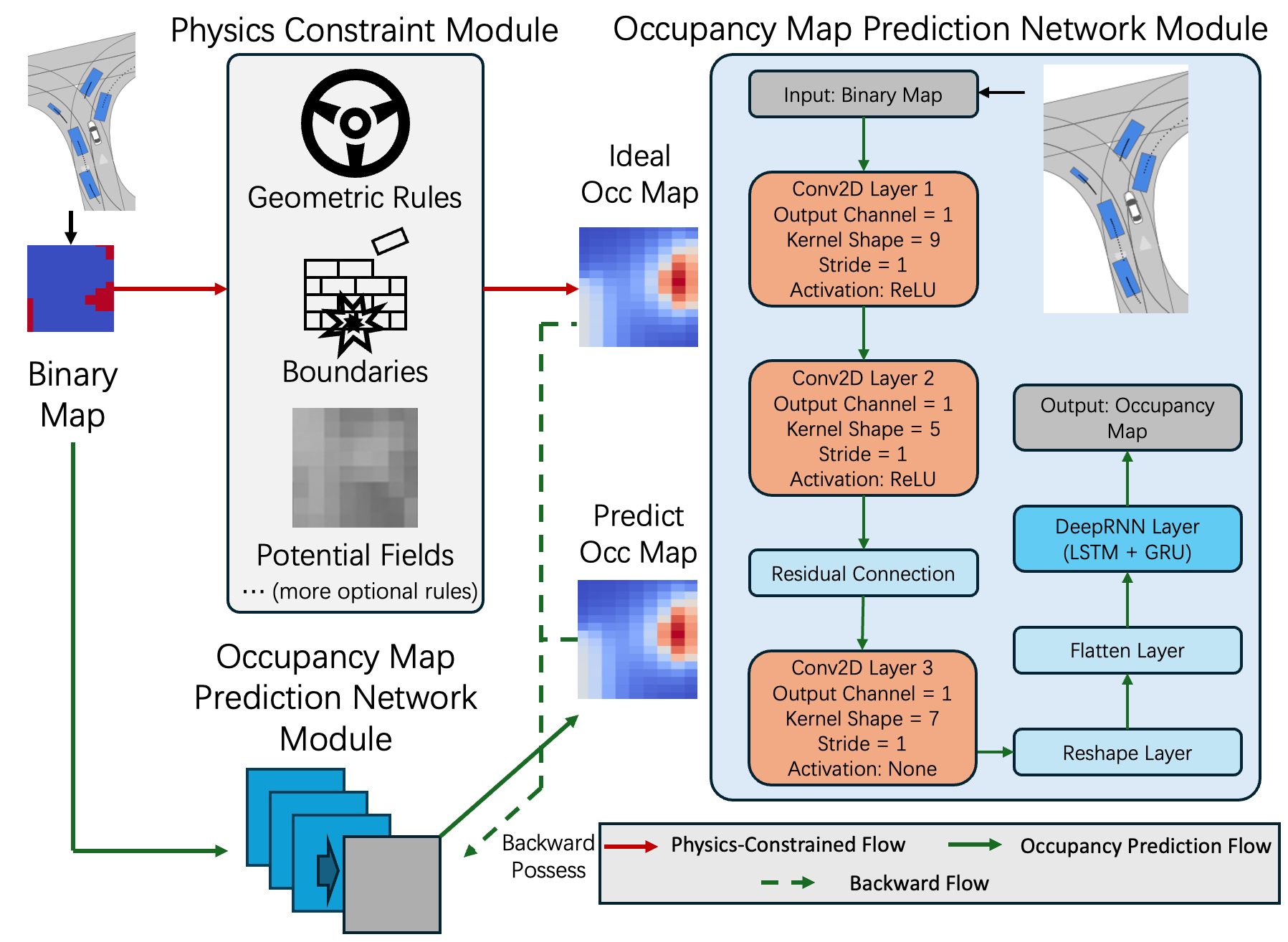}
    \caption{The framework combines a physics constraint module and a prediction network. The constraint module generates ideal occupancy maps using geometric rules, boundaries, and potential fields. The prediction network takes binary maps as input and outputs predicted occupancy maps via convolutional and recurrent layers. Learning is guided by the difference between predicted and ideal maps.}
    \label{framework}
\end{figure*}

\subsection{Network Architecture for End-to-End Occupancy} 
The architecture of our end-to-end occupancy map prediction network is designed to efficiently integrate input data and generate predicted occupancy maps through multi-level processing. As illustrated, the system consists of a series of convolutional layers, with residual connections employed to maintain feature map consistency and ensure that information is preserved as it flows through the network.

As shown in Fig. \ref{framework}, the workflow of the network begins with the input of a binary map representing the driving environment or scene data, which is then processed through the network.
The subsequent convolutional layers are responsible for extracting features from the input data. The first convolutional layer has one output channel with a kernel size of 9 and applies the ReLU activation function to introduce nonlinearity, enabling the network to capture complex patterns in the input:
\begin{equation}
F_1 = \text{ReLU}(W_1 * I + b_1)
\end{equation}
where $F_1$ is the output feature map of the first convolutional layer, $W_1$ is the convolution kernel, $I$ is the binary input map, and $b_1$ is the bias term.
The second convolutional layer uses a kernel size of 5 and also employs the ReLU activation function:
\begin{equation}
F_2 = \text{ReLU}(W_2 * F_1 + b_2)
\end{equation}
where $F_2$ is the output of the second convolutional layer, $W_2$ is the second layer's kernel, and $b_2$ is the corresponding bias.
Residual connections are then incorporated to ensure that critical feature information is preserved:
\begin{equation}
F_{\text{res}} = F_2 + F_1
\end{equation}
where $F_{\text{res}}$ is the result of the residual connection combining $F_1$ and $F_2$.
The third convolutional layer that follows has an output channel of 1 and a kernel size of 7, with no activation function applied, allowing information to be passed through more directly:
\begin{equation}
F_3 = W_3 * F_{\text{res}} + b_3
\end{equation}
where $F_3$ is the output of the third convolutional layer, $W_3$ is the convolution kernel, and $b_3$ is the bias term.

The network then reshapes the output into a format suitable for processing by recurrent layers, followed by a flattening layer that prepares the data for further computation. A deep RNN layer, combining LSTM and Gated Recurrent Unit (GRU) architectures, is employed to handle temporal sequences and capture temporal dependencies within the occupancy maps—an essential capability for prediction in dynamic environments:
\begin{equation}
Z_t = \text{GRU}(\text{LSTM}(F_3, h_{t-1}), h_t)
\end{equation}
where $Z_t$ is the output of the GRU at time step $t$, $F_3$ is the spatial feature input, $h_{t-1}$ and $h_t$ are the hidden states of the LSTM and GRU layers, respectively.

Finally, the network outputs an occupancy map with a shape of $[B, 36, 9]$, where $B$ denotes the batch size, and 36 and 9 represent the height and width of the occupancy map, respectively.

The architecture is designed with the dynamic nature of autonomous driving environments in mind, enabling the network to capture the temporal evolution of obstacles and vehicle positions. By incorporating recurrent layers, the network can effectively model changes over time, ensuring accurate occupancy map predictions.

\subsection{Physics-informed Constraint via Artificial Potential Fields} 
In this study, we enhance the accuracy of occupancy map prediction by incorporating physical rule constraints. These constraints ensure that the model's predictions align with real-world physical principles—particularly in maintaining safe distances—thereby effectively preventing collisions between vehicles and obstacles.

Physical rule constraints can be implemented using various physical principles; in this work, we adopt Artificial Potential Fields (APF) as the chosen constraint. The core idea of APF is to incorporate geometric rules that ensure the boundaries of the occupancy map conform to real-world road geometries, traffic boundaries, and lane markings.

Additionally, the potential field functions by applying attractive forces to guide the vehicle toward the target area and repulsive forces to prevent collisions with obstacles. The total potential at a position $\mathbf{x}$ is defined as:
\begin{equation}
U(\mathbf{x}) = U_{\text{att}}(\mathbf{x}) + U_{\text{rep}}(\mathbf{x})
\end{equation}
where $U(\mathbf{x})$ is the total potential at position $\mathbf{x}$, $U_{\text{att}}$ is the attractive potential, and $U_{\text{rep}}$ is the repulsive potential.

\begin{equation}
U_{\text{att}}(\mathbf{x}) = \frac{1}{2} \xi \|\mathbf{x} - \mathbf{x}_{\text{goal}}\|^2
\end{equation}
where $\xi$ is a positive scaling factor for attraction, $\mathbf{x}_{\text{goal}}$ is the target position, and $\|\cdot\|$ denotes the Euclidean norm.

\begin{equation}
U_{\text{rep}}(\mathbf{x}) =
\begin{cases}
\frac{1}{2} \eta \left( \frac{1}{\|\mathbf{x} - \mathbf{x}_{\text{obs}}\|} - \frac{1}{d_0} \right)^2, & \text{if } \|\mathbf{x} - \mathbf{x}_{\text{obs}}\| \leq d_0 \\
0, & \text{otherwise}
\end{cases}
\end{equation}
where $\eta$ is the repulsion scaling factor, $\mathbf{x}_{\text{obs}}$ is the position of the obstacle, and $d_0$ is the repulsion influence distance threshold.

The ideal occupancy map is generated by incorporating these physical rules, and during learning process, the predicted occupancy map is compared against the ideal one. The network learns by minimizing the discrepancy:
\begin{equation}
\mathcal{L} = \frac{1}{HW} \sum_{i=1}^{H} \sum_{j=1}^{W} \left( \hat{O}_{i,j} - O^*_{i,j} \right)^2
\end{equation}
where $\hat{O}_{i,j}$ is the predicted occupancy value at cell $(i,j)$, $O^*_{i,j}$ is the ideal (ground truth) occupancy value, and $H$, $W$ denote the height and width of the occupancy map.

It is worth noting that although this study employs Artificial Potential Fields (APF) as the physical constraint, the approach is highly extensible. Other forms of physical rules can be substituted for APF without modifying the network architecture, allowing the framework to adapt to different task requirements or physical priors.

In summary, the proposed method integrates an end-to-end deep learning network with physical rule constraints to improve the accuracy of occupancy map prediction. The network architecture extracts features through convolutional layers and captures temporal dependencies via recurrent layers, while the incorporation of physical rules ensures the physical plausibility of the predictions, enhancing safety, controllability, and adaptability. This approach effectively improves the reliability and interpretability of autonomous driving systems, providing essential technical support for future applications.

\section{Experimental Result}
In this section, we provide a comprehensive presentation and analysis of the experimental results from three perspectives: model training, representative scenario analysis, and overall evaluation. First, in the model training part, we illustrate the variation of the loss function during learning process to verify the model’s convergence and stability. Next, we select two representative traffic interaction scenarios to compare the trajectories generated by our model with the ground truth, visually demonstrating the model’s ability to predict behavior under different conditions. Finally, in the overall evaluation, we report the model’s performance across all test scenarios using key metrics such as Time to Collision (TTC)\cite{ortiz2023road} and task completion rate \cite{sharath2021literature}. These results are summarized in tabular form to provide a thorough assessment of the effectiveness and advantages of the proposed method.

\subsection{Model Training}
\begin{figure}[t]
    \centering
    \includegraphics[width=0.5\textwidth]{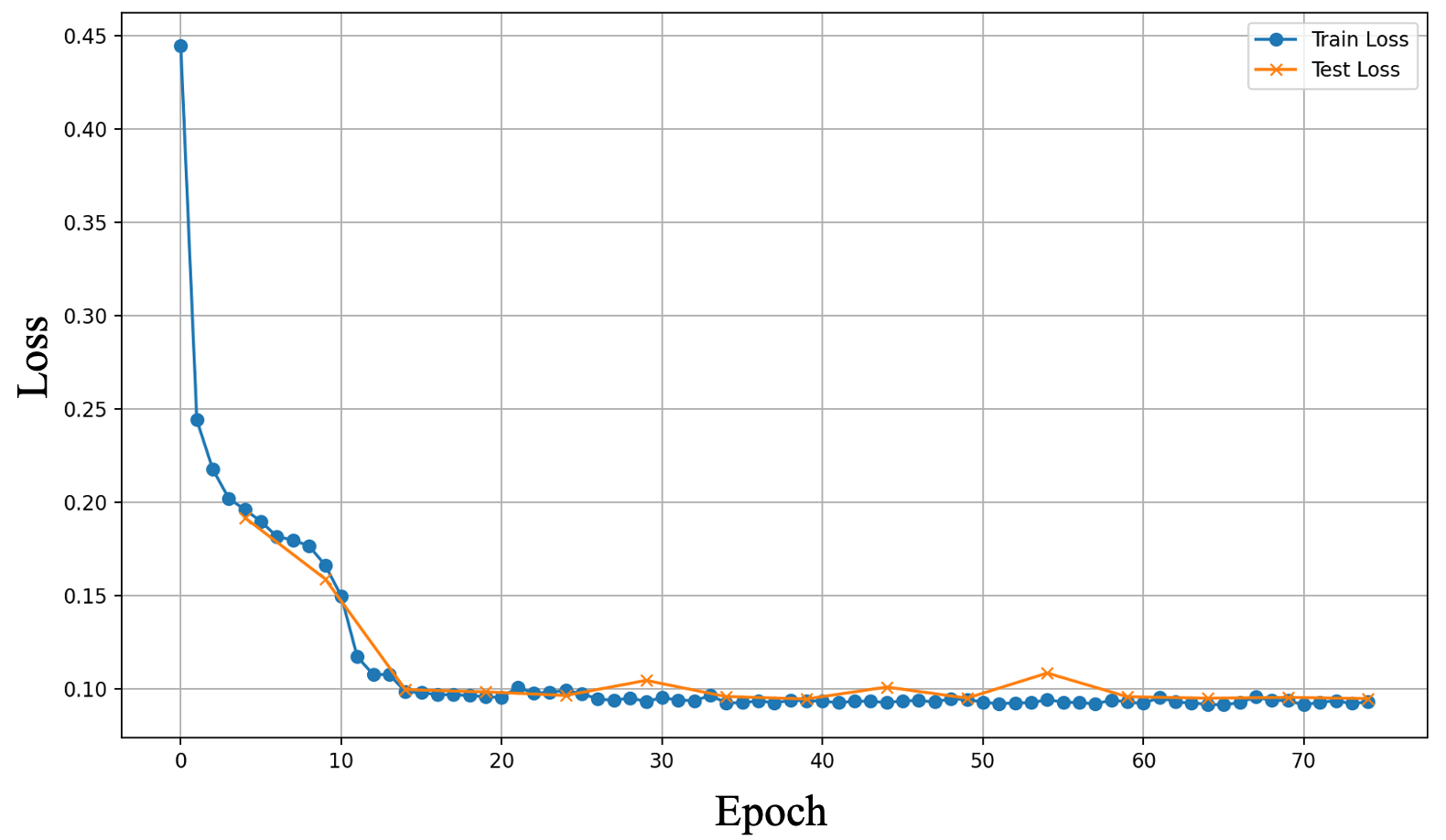}
    \caption{This figure illustrates the training and testing loss curves of the proposed model during the learning process. The x-axis represents the number of epochs, while the y-axis denotes the loss values. The blue solid line indicates the training loss, which decreases rapidly in the initial epochs and then stabilizes around a low value. The orange dashed line represents the testing loss, showing a similar trend but with slight fluctuations. The convergence of both curves suggests that the model has achieved good generalization performance without significant overfitting.}
    \label{loss}
\end{figure}
In this section, we evaluate the convergence and training stability of the model by visualizing the loss trend during the learning process. The loss is defined as the sum of all elements in the difference between the potential field map generated by the physical rule and that generated by the network. Fig. \ref{loss} shows the loss curves on both the training and validation sets, where the x-axis represents the total number of training iterations, and the y-axis indicates the average loss computed after each iteration.

Overall, the training loss decreases rapidly in the early stages, indicating that the model quickly learns the fundamental patterns and rules in the data. As training progresses, the rate of loss reduction gradually slows down and stabilizes around the 14th epoch, suggesting that the model has reached a good convergence state. Meanwhile, the loss trend on the validation set closely follows that of the training set, with no signs of significant overfitting or divergence, further confirming the effectiveness of the model architecture and training strategy.

In summary, the systematic analysis of the loss curves during training confirms that the proposed model can accurately mimic the potential field maps generated by physical rules, providing a reliable foundation for subsequent behavior prediction and trajectory generation tasks.

\subsection{Typical Scenario}
\begin{figure}[t]
    \centering
    \includegraphics[width=0.5\textwidth]{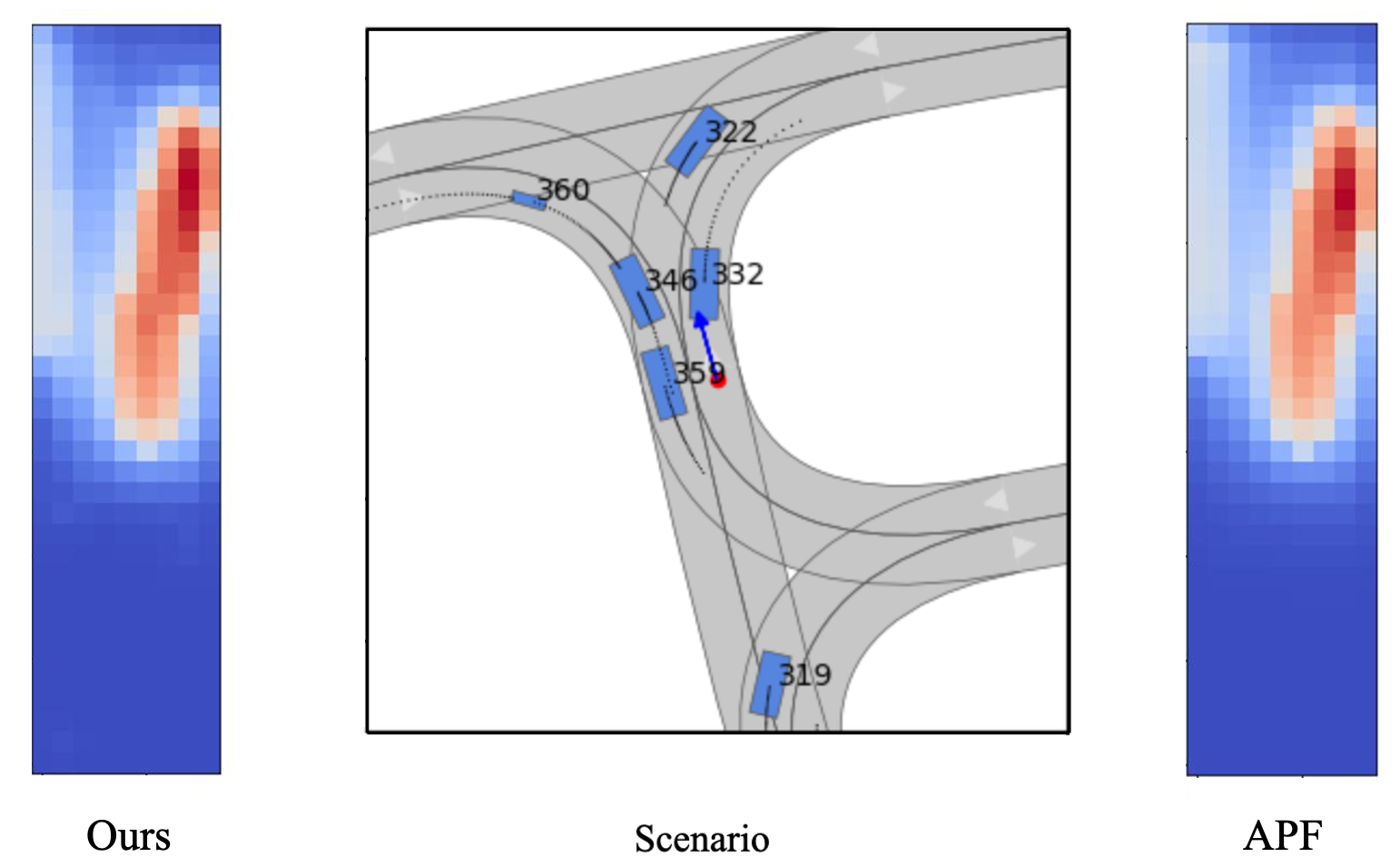}
    \caption{The target vehicle (red arrow) interacts with surrounding vehicles (blue boxes). The network-based method (left) generates a smoother and more adaptive potential field, while the rule-based method (right) shows a more rigid distribution. Red indicates high-risk areas; blue indicates low-risk regions.}
    \label{typical}
\end{figure}
To further validate the modeling capability of the proposed method in complex traffic scenarios, we conducted a comparative analysis between the APF method and ours by examining the potential field maps generated under the same scenario, as shown in Fig. \ref{typical}. The scenario includes a target vehicle and several surrounding dynamic traffic participants. Red regions indicate high potential energy (i.e., potential conflict zones), while blue regions represent low potential energy (i.e., relatively safe areas).

As observed from the visualization results, the potential field map generated by our method exhibits a more refined spatial distribution. Unlike the APF method, where high-potential regions remain consistent, our method assigns higher potential values to areas closer to the ego vehicle, accurately capturing the interaction risks between the target vehicle and surrounding vehicles. In contrast, although the APF method can also identify major risk zones, its field distribution is more rigid and lacks responsiveness to dynamic behavioral changes, failing to adequately represent complex interaction relationships.

Notably, the network-based method, through its end-to-end learning mechanism, can automatically extract scene features and effectively model the interaction effects between vehicles, demonstrating stronger environmental adaptability and a better ability to capture uncertainty. The above comparison results indicate that, compared to traditional manually designed rule-based approaches, the proposed method not only achieves higher prediction accuracy but also exhibits superior generalization and learning capabilities, offering significant advantages in behavior prediction tasks under complex traffic conditions.

\subsection{Two Sample Cases}
\begin{figure*}[t]
    \centering
    \includegraphics[width=0.6\textwidth]{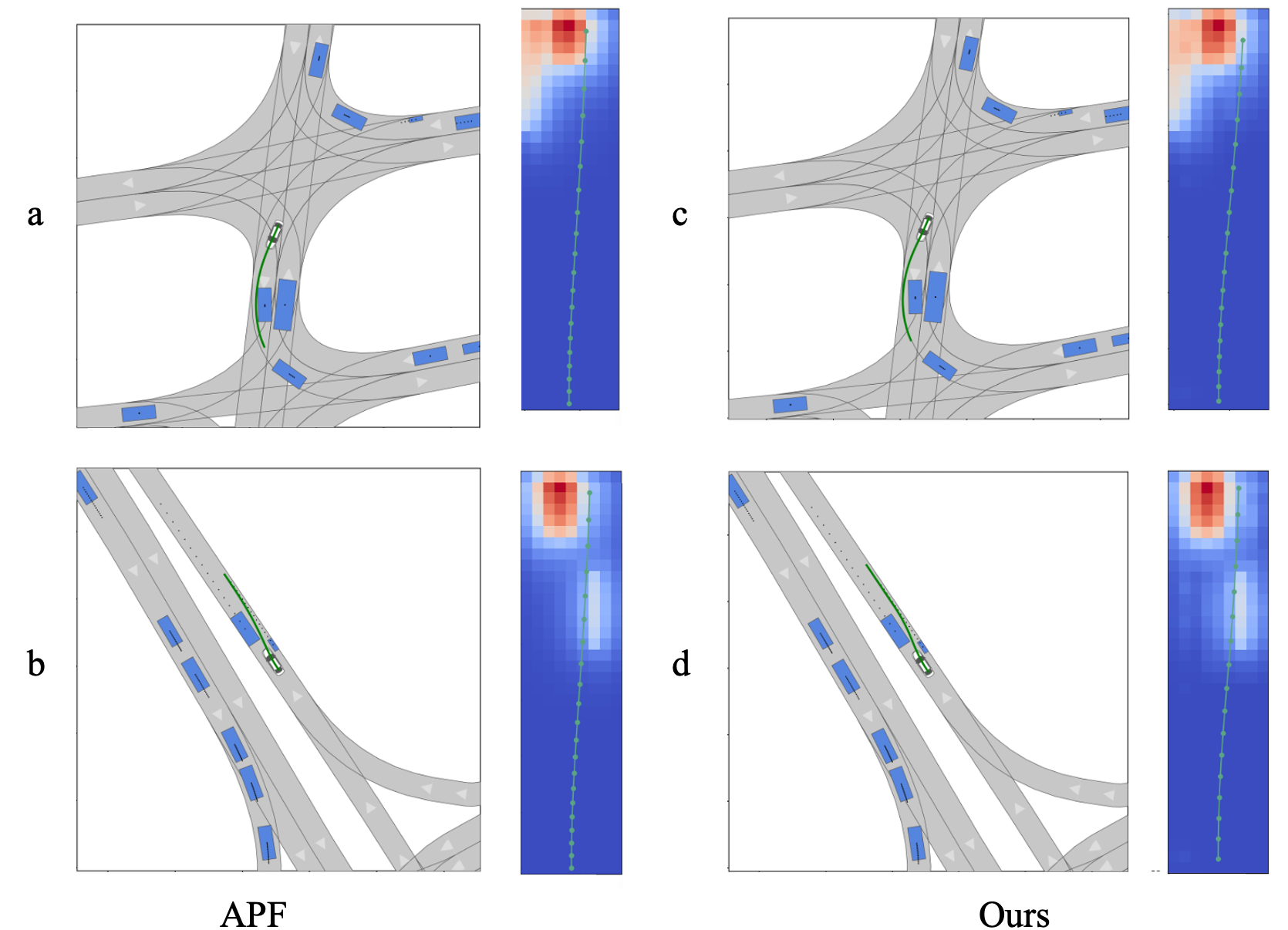}
    \caption{This figure compares trajectory predictions of APF method and Ours in two complex traffic scenarios. The green trajectory represents the target vehicle, while the blue boxes indicate surrounding vehicles. Heatmaps on the right show potential high collision risk. Ours generates more flexible and realistic trajectories compared to the smoother but less adaptive paths produced by the APF method.}
    \label{2cases}
\end{figure*}
To intuitively assess the behavior prediction capability of the proposed method in complex traffic environments, we selected two representative interactive scenarios for qualitative analysis, with the visualization results shown in Fig. \ref{2cases}. In the figure, the green trajectory represents the predicted path of the target vehicle, blue dots indicate the positions of surrounding traffic participants, red regions highlight areas of potential high collision risk, and the heatmap reflects the model's probabilistic prediction of the target vehicle’s future trajectory.

In the first scenario, as shown in Fig. \ref{2cases}a, \ref{2cases}c, ours enables the target vehicle to more effectively avoid high-risk areas based on the generated potential field map compared to the APF method, while the resulting trajectory aligns more closely with real-world driving behavior.
In the second scenario, as shown in Fig. \ref{2cases}b, \ref{2cases}d, the trajectory generated by APF, while overall smooth, fails to adequately adapt to the dynamic behavior of surrounding vehicles, resulting in a relatively higher risk of passing through high-risk areas. In contrast, ours demonstrates stronger environmental awareness and greater flexibility in behavior prediction.

In summary, the comparison of the two methods in representative traffic scenarios reveals that APF method face limitations in handling dynamic interactions and uncertainty. In contrast, the proposed method behavior prediction method demonstrates superior capability and robustness in terms of trajectory plausibility, interaction adaptability, and safety risk identification.

\subsection{Holistic Evaluation}

\begin{table}[]
\caption{Comparison Results of APF and Ours}
\renewcommand{\arraystretch}{1.5} 
\setlength{\tabcolsep}{9pt}     
\begin{tabular}{cccccc}
\hline
\multicolumn{6}{c}{\textbf{Results Among 2000 Commonroad Scenarios}} \\ \hline \hline
\textbf{Method}  & \textbf{TCR} & \textbf{TTC} & \textbf{Jerk} & \textbf{Head-Way} & \textbf{Time} \\ \hline
\textbf{APF}   & 0.902          & 2.798          & 2.079          & 19.054          & 0.01            \\ \hline
\textbf{Ours} & \textbf{0.946} & \textbf{2.979} & \textbf{1.361} & \textbf{21.124} & \textbf{0.0019} \\ \hline
\end{tabular}
\label{tabel1}
\end{table}

To comprehensively evaluate the overall performance of the proposed method in complex traffic environments, we conducted experiments on 2,000 CommonRoad scenarios and benchmarked it against the rule-based method (CommonRoad-Reactive-Planner). Table~\ref{tabel1} summarizes the performance comparison across several key metrics, including Task Completion Rate \cite{sharath2021literature}, Time-to-Collision (TTC) \cite{ortiz2023road}, Jerk \cite{leejerk}, Headway \cite{parashar2025reassessing}, and Execution Time.

Experimental results demonstrate that our method achieved a task completion rate of 0.946, higher than the baseline’s 0.902, indicating a stronger capacity to generate safe and feasible trajectories in diverse interactive scenarios. Regarding TTC, our approach yielded a longer average time of 2.979 seconds compared to 2.798 seconds by the baseline, reflecting improved safety through earlier collision avoidance.
In terms of Jerk, our method significantly reduced the average value to 1.361 m/s³, compared to 2.079 m/s³ by the baseline, suggesting a smoother and more comfortable driving experience. Similarly, the Headway metric was improved, with our method maintaining an average spacing of 21.124 meters versus the baseline’s 19.054 meters, further enhancing safety margins.
Most notably, our approach demonstrated a substantial improvement in computational efficiency. The average execution time was reduced to only 0.0019 seconds, which is over five times faster than the baseline’s 0.01 seconds. This drastic reduction in planning time not only enables real-time deployment but also provides ample margin for replanning in dynamic environments, highlighting the practical value of the proposed method in time-critical autonomous driving systems.

In summary, our method delivers superior performance across success rate, safety, smoothness, and efficiency, making it highly suitable for real-world autonomous driving applications.

\section{Conclusion}



In this work, we present a unified end-to-end motion planning framework that integrates physical constraints into the occupancy prediction process. By embedding APF as physics-informed priors, the predicted occupancy maps respect both data-driven learning and real-world physical rules. This improves the safety, interpretability, and generalization of the model. Our framework effectively combines convolutional and recurrent neural networks with explicit physical guidance, offering a robust and adaptive solution for autonomous vehicle motion planning.

Through experiments, we show that introducing physical constraints during training enhances both safety and computational efficiency. The framework remains highly flexible, allowing different physical rules to be incorporated without changing the network architecture, thus enabling broad applicability across various driving tasks.

Future work will aim to optimize the framework further, explore alternative forms of physical constraints, and extend its applicability to more complex scenarios with higher uncertainty.

\section*{Acknowledgment}
This study is supported by RGC General Research Fund (GRF) HKUST16205224, NSFC Grant U24A20252 and 62373315, and the Red Bird MPhil Program at the Hong Kong University of Science and Technology (Guangzhou)

\bibliographystyle{IEEEtran}
\bibliography{main}

\end{document}